\documentclass{article}

\usepackage{arxiv}

\usepackage[utf8]{inputenc} 
\usepackage[T1]{fontenc}    
\usepackage{hyperref}       
\usepackage{url}            
\usepackage{booktabs}       
\usepackage{amsfonts}       
\usepackage{nicefrac}       
\usepackage{microtype}      
\usepackage{lipsum}		
\usepackage{graphicx}
\usepackage{natbib}
\usepackage{doi}
\usepackage{subcaption}

\title{NLP and Education: \\ using semantic similarity to evaluate filled gaps \\ in a large-scale Cloze test in the classroom}

\author{ \href{https://orcid.org/0009-0000-5270-8033}{\includegraphics[scale=0.06]{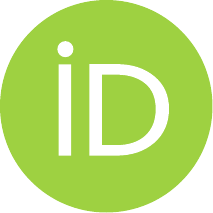}\hspace{1mm}Túlio Sousa de Gois} \\
	Departamento de Computação \\
    Universidade Federal de Sergipe - UFS\\
	\texttt{tuliosg@academico.ufs.br} \\ 
	\And
	\href{https://orcid.org/0009-0008-1795-6822}{\includegraphics[scale=0.06]{orcid.pdf}\hspace{1mm}Flávia Oliveira Freitas} \\
	Departamento de Letras Vernáculas \\
    Universidade Federal de Sergipe - UFS\\
	\texttt{fla5882freitas@academico.ufs.br} \\
    \And
    \href{https://orcid.org/0000-0003-0275-3578}{\includegraphics[scale=0.06]{orcid.pdf}\hspace{1mm}Julian Tejada} \\
	Departamento de Psicologia \\
    Universidade Federal de Sergipe - UFS\\
	\texttt{jtejada@academico.ufs.br} \\
    \And
    \href{https://orcid.org/0000-0002-4972-4320}{\includegraphics[scale=0.06]{orcid.pdf}\hspace{1mm}Raquel Meister Ko. Freitag}\\
	Departamento de Letras Vernáculas \\ Universidade Federal de Sergipe -  UFS\\
	\texttt{rkofreitag@academico.ufs.br} \\
}

\date{}

\renewcommand{\headeright}{}
\renewcommand{\shorttitle}{NLP and Education: using semantic similarity to evaluate filled gaps in a large-scale Cloze test in the classroom}

\begin{document}
\maketitle

\begin{abstract}
    This study examines the applicability of the Cloze test, a widely used tool for assessing text comprehension proficiency, while highlighting its challenges in large-scale implementation. To address these limitations, an automated correction approach was proposed, utilizing Natural Language Processing (NLP) techniques, particularly word embeddings (WE) models, to assess semantic similarity between expected and provided answers. Using data from Cloze tests administered to students in Brazil, WE models for Brazilian Portuguese (PT-BR) were employed to measure the semantic similarity of the responses. The results were validated through an experimental setup involving twelve judges who classified the students' answers. A comparative analysis between the WE models' scores and the judges' evaluations revealed that GloVe was the most effective model, demonstrating the highest correlation with the judges' assessments. This study underscores the utility of WE models in evaluating semantic similarity and their potential to enhance large-scale Cloze test assessments. Furthermore, it contributes to educational assessment methodologies by offering a more efficient approach to evaluating reading proficiency. 
\end{abstract}


\section{Introduction}

Since half past the last century, the Cloze test has been used for educational purposes to assess proficiency in understanding texts in different languages \cite{taylor1953cloze, brown1980relative, brown2002cloze}. The task consists of the systematic filling in of gaps in a text, specifically a prose selection \cite{bickley1970cloze}, previously adapted to the participant’s realities, and the scores of correct answers are associated with the degree of comprehension of the text by the participant.

Different measures, such as exact answer, acceptable answer \cite{brown1980relative}, multiple choice, and Clozentropy \cite{darnell1968development, lowry1975clozentropy}, have been used to assess gap-filling since Taylor's initial proposal \cite{taylor1953cloze}. These measures will be further examined in Section 2. The exact answer may seem easier to calculate, especially for a Cloze test applied to large and heterogeneous groups of students with insufficient time for teachers to analyze each answer individually. In Brazil, for instance, teachers usually have to manage numerous classes, and this correction method helps to provide rapid answers to students’ reading proficiency, allowing one to check the answers objectively \cite{cunha2010estudos} without possible or different options. 

The exact answer is the most common measure adopted in large-scale studies and classroom activities due to the easier procedure to correct the gaps filled. However it is not the best way to measure comprehension since reading entails top-down and bottom-up reading strategies, and filling the gaps with a unique and specific word from the original text limits the reading approach underlying the Cloze test by avoiding other answer possibilities, like words with similar meanings like synonyms.  

An automatic correction of large-scale assessment that evaluates reading proficiency provides a faster and more accurate evaluation of responses. Techniques from Natural Language Processing (NLP) stand out among the ways of automating this task, as they provide the computer with forms of word representation and allow it to carry out different calculations using these representations. In this work, we suggest extending the measure by calculating the semantic similarity between the answer given and the correct one, using word embeddings (WE) models. 
    
\section{The Cloze procedure} \label{sec:firstpage}
\begin{figure}[!ht]
    \centering
    \includegraphics[width=0.5\linewidth]{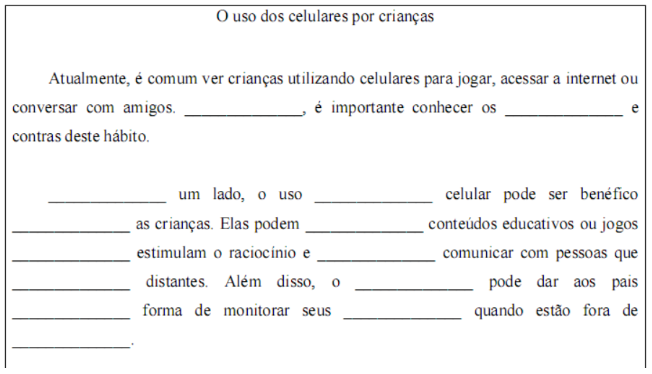}
    \caption{The text “The use of cellphones by children”, in Pt-PB was developed for large-scale diagnostic use (\cite{cardoso2024eficiencia}).}
    \label{fig:cloze test}
\end{figure}

The Cloze test considers the reader’s cognitive aspects, measures integrative skills, and implies that the gaps are contextually interrelated. The task consists of creating or adapting a prose text of appropriate difficulty and systematically deleting every nth word supplied by the reader \cite{bickley1970cloze, brown1980relative, taylor1953cloze, taylor1957cloze, oller1971cloze}. The set of the Cloze test must fulfill the objective of its developer and be appropriate to the reader’s age, and school level \cite{oller1971cloze}, for instance, to promote a better assessment quality.  Figure \ref{fig:cloze test}.

\cite{brown1980relative} presents four kinds of Cloze scoring methods: the exact answer (EX), the acceptable answer (AC), multiple-choice and Clozentropy. The EX method counts only the word deleted from the original passage, classifying the answers as right /wrong \cite{lowry1975clozentropy}. The AC “usually counts any contextually acceptable answer as correct” \cite{brown1980relative}. Multiple-choice method offers the test taker alternative answers from which the subject has to choose the correct one for each blank \cite{brown1980relative}. And Clozentropy scoring method compares each subject’s response to all of the answers provided by some specific group \cite{lowry1975clozentropy}. The way in which gaps are assessed underlies a view of reading. 

The measures of the exact answer (EX) and the acceptable answer (AC) presuppose reading as a process restricted to the domain of the text. They are easier to use to calculate proficiency. However, if we understand reading as the interaction of the speaker's knowledge with the text, the Clozentropy method is more appropriate. This type of measure requires more than a list of items to be checked, which is why a computational approach can be implemented. One of Clozentropy’s assumptions is that language norms vary from group to group, and the best measure of proficiency is in terms of group or groups with whom the subject needs to communicate \cite{lowry1975clozentropy}. In this scoring method, the answers are not compared to the original passage, instead, they are considered according to the members of a criterion group who have taken the same test. Depending on the difference between the criterion group and the test takers, it may represent a heterogeneous educational level and be more complex to assess.

\section{Semantic Similarity}
Identified as the measurement of semantic similarity between two texts, semantic textual similarity (STS) seeks to establish a percentage of similarity or a ranking between these texts \cite{chandrasekaran2021evolution}. In the Natural Language Processing field, it is one of the most challenging and open research problems, resulting in different approaches and methods emerging to tackle the problem over the years \cite{chandrasekaran2021evolution, mohammad2012distributional}. 

One of the most efficient ways of representing words is through word embeddings (WE), numerical vectors representing a word in an n-dimensional space \cite{hartmann2017portuguese}. These vectors can preserve the linguistic relationships between words and be generated via various models like word2vec \cite{mikolov2013efficient} and GloVe \cite{pennington2014glove}. Each of these models employs a distinct algorithm to generate word embeddings, and they can be categorized into two primary groups: those that leverage co-occurrence matrix, such as GloVe, and those that are predictive models, aiming to predict a word based on its context, such as Word2Vec \cite{hartmann2017portuguese}.

By exploiting the distributional hypothesis (based on the idea that similar words occur together frequently) in constructing vectors, WEs can be classified under corpus-based semantic-similarity methods \cite{gorman2006scaling}. As different models have emerged for constructing vectors for text representation, diverse ways of measuring the semantic similarity between them have also been proposed. Despite all these measures, cosine similarity has gained prominence and its use has become more frequent in NLP research \cite{mohammad2012distributional}. 

To conduct this study and find the best scoring method, we applied a Cloze test entitled “The use of cellphones by children” to students of an elementary public school in Sergipe, Brazil.  To evaluate the children’s test performance, we used WEs for Brazilian Portuguese generated by three different models: GloVe, Wang2Vec \cite{ling2015two}, and spaCy; and calculated semantic similarity using cosine similarity.

\section{Method}
An experimental study was performed to evaluate the usability of WEs as a method for proofreading the Cloze test, comparing the classification of the answers of a test application completed by a group of judges and three models of WEs.

\subsection{Participants}
24 sixth graders (M age  = 12,7) from an elementary public school in Sergipe, Brazil, participated in the data collection. In addition, twelve post-graduate students from the post-graduate courses in Linguistics and Psychology of the Federal University of Sergipe participated as judges of the student’s responses.

\subsection{Cloze test procedure}
A specific 200-word text was created by the AI ChatGPT3.5. The title and the initial sixteen words remained the same as the original passage, followed by the gap at each five words. The students filled the gaps in a paper test. Responses were tabulated in a spreadsheet.  

\subsection{Assessment of similarity by humans}
The tabulated responses for each gap were distributed to the judges who were asked to rank-order them from most to least appropriate, by filling in an online form built with surveyjs (\verb|https://github.com/surveyjs/survey-library|) and distributed using JATOS \cite{lange2015just}. To classify the answers, the judges checked the sentence with the gap and all the answers provided and then sorted them. The instruction to create the ranking consisted of ordering the words so that those higher up made more sense in the context of the given sentence.

\subsection{Assessment of similarity by Word Embeddings models}
In the context of Brazilian Portuguese, \cite{hartmann2017portuguese} present different WEs models trained on a corpus of Portuguese (including both Brazilian and Portuguese variants) and their evaluations in tasks such as POS tagging, sentence semantic similarity, and semantic analogies.We selected the models that exhibited the best performance on sentence semantic similarity and semantic analogy tasks: GloVe 600-dimensional, which excelled in semantic analogies, and Wang2Vec 1000-dimensional, which outperformed others in sentence semantic similarity.

Another model used for semantic similarity is pt\_core\_news\_lg\footnote{https://spacy.io/models/pt\#pt\_core\_news\_lg}, a pipeline for spaCy that provides word vectors with 300 dimensions. As a library with various tools and models for NLP, spaCy makes it easy to apply methods and carry out similarity measurements. For this reason, the student’s answers were submitted to the semantic similarity evaluation using the selected WE models, which calculated the use of the cosine similarity between each answer given and the one expected. Cosine similarity was chosen as a metric due to its popularity and effectiveness in measuring the similarity between vectors in high-dimensional spaces, a common approach in NLP research \cite{chandrasekaran2021evolution}.

\subsection{Validation}
To choose the appropriate WE model for evaluating Cloze test responses, we developed an experiment in which the models and twelve judges assessed the answers, comparing them.

From the spreadsheet, the gaps with more than 10 alternatives were selected for our initial analysis, leaving 19 of the 37 gaps with around 159 words. The judges classified the children's answers according to each sentence with a gap. Based on the classification obtained for each gap, the first position words were removed, leaving the remaining 154 words.
In the end, the outputs from each model were ordered in descending order in terms of similarity value to be compared with the judges' rankings using non-parametric correlation.

\section{Results}
The results obtained from the experiment with the judges and the outputs of the WE models were compared and subjected to different analyses. Examining the heterogenous rankings  (judges and WE’s models) through the Analysis of Variance of Aligned Rank Transformed Data \cite{wobbrock2011aligned}, no significant difference was found, $(F(3,717)=0.3486, 
 p=0.7917)$. 

\begin{figure}[!ht]
    \centering
    \includegraphics[width=0.5\linewidth]{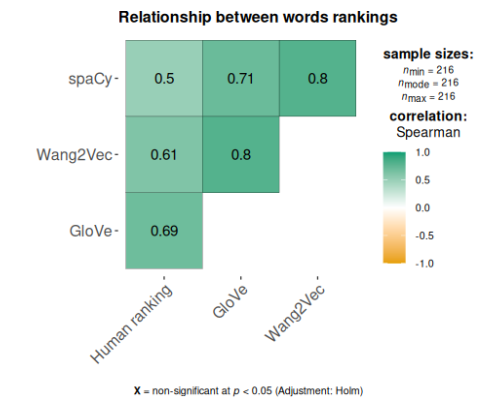}
    \caption{Relationship between words rankings}
    \label{fig:ranking}
\end{figure}

The nonparametric Spearman correlations showed significant high correlations between the different ranks, with GloVe as the best WE model. GloVe showed the highest correlation when comparing the ranking classification of all words (see Figure \ref{fig:ranking}).

\section{Conclusion}
NLP methods can help with the limitations in Cloze test evaluation. Using WE models and the calculation of semantic similarity, in a case study validates these techniques as a new way of evaluating tests. We performed the validation by creating different rankings in a sorting task accomplished by humans and compared them with the outputs generated by the models.

Through the validation of the experiment, the model that performed best in the analysis was GloVe, showing the highest correlation both between the other models and with the judges' ranking, corroborating the \cite{hartmann2017portuguese} results in terms of performance in semantic similarity tasks. Based on the high correlation between the ratings and the absence of significant differences between the ratings generated by the models and those constructed by the judges, it is clear that WE models are useful for automating the correction of cloze tests using semantic similarity. 

Furthermore, it is important to note that the WE models presented here are based on training on large corpora, making them corpus-based models. As NLP has advanced, deep network-based–methods that use WEs to transform text data into high-dimensional vectors have emerged, and the efficiency of these embeddings improves performance in tasks involving semantic similarity \cite{levy2014neural}. Consequently, future work is suggested to use deep network-based methods to create embeddings, evaluate their performance in the semantic similarity task to correct the Cloze test, and compare the results with corpus-based methods.

\paragraph{Acknowledgments.}
We would like to express our sincere gratitude to the infrastructure and staff of the Multiuser Laboratory for Informatics and Linguistic Documentation (LAMID) for their invaluable support throughout this research. We are particularly grateful to the team of administrators who conducted the cloze tests, whose dedication was instrumental in the data collection process.
We acknowledge the financial support of the National Council for Scientific and Technological Development (CNPq).

\bibliographystyle{unsrtnat}
\bibliography{references}

\end{document}